\documentclass[letterpaper,journal]{IEEEtran}
\usepackage{amsmath,amsfonts}
\usepackage{algorithm}
\usepackage[noend]{algpseudocode}
\usepackage{amsthm}
\usepackage{graphicx}
\usepackage{hyperref}

\newtheorem{theorem}{Theorem}
\newtheorem{definition}[theorem]{Definition}

\begin{document}

\title{Achieving Asymptotic Near-Optimality Without Delta-Similarity}

\author{Michael Moncton and Eric Frew}




\maketitle

\begin{abstract}
Sampling-based motion planning algorithms are a popular class of trajectory planning algorithm due to their speed in complex, high-dimensional environments and ability to handle kinodynamic constraints, specifically through the use of forward dynamics propagation.
Many such planners claim to achieve asymptotic near-optimality by proving the almost sure sampling of trajectories that are close to an optimal trajectory in the state space, known as $\delta$-similar trajectories.
This paper shows that the proof behind asymptotic $\delta$-similarity relies on an unstated assumption that $\delta$-similar trajectory segments will always be kept once sampled.
This assumption does not hold in general.
A problematic case, referred to as ``crowding out,'' is described, where locally low-cost paths prevent trajectories that are $\delta$-similar to the optimal trajectory from being added to the tree.
It is shown, however, that asymptotic near-optimality guarantees can still be achieved without guarantees of $\delta$-similar solution trajectories when crowding out is properly accounted for.
An example environment and system are provided where crowding out is shown to occur, demonstrating a scenario where inductively sampling a $\delta$-similar solution trajectory is impossible.
\end{abstract}

\begin{IEEEkeywords}
Motion planning, kinodynamic planning, sampling-based algorithms, asymptotic near-optimality, asymptotic optimality
\end{IEEEkeywords}

\section{Introduction}
Kinodynamic sampling-based motion planning (SBMP) algorithms have been a mainstay in robotics because of their ability to find feasible paths quickly in high-dimensional environments with complex dynamic constraints \cite{LaValle2001RandomizedPlanning, Kavraki1996ProbabilisticSpaces, Hsu1997PathSpaces}.
Some kinodynamic SBMP algorithms are able to converge towards optimal solution paths over time, a property called asymptotic optimality \cite{Karaman2011Sampling-basedPlanning, Shome2021AsymptoticallyEdges, Hauser2016AsymptoticallySpace}.
Asymptotically near-optimal (ANO) SBMP algorithms converge similarly, but are only guaranteed to reach costs within a bounded range of an optimal solution \cite{Webb2013KinodynamicDynamics, Dobson2014SparsePlanning}.
To generate such near-optimal trajectories quickly, dynamics can be forward-propagated to generate a trajectory tree, which is then pruned to maintain a tree that increases in quality over time, such as in the Stable Sparse RRT (SST) algorithm \cite{Li2016AsymptoticallyPlanning}.
In addition to improving the quality of the trajectory tree, the pruning process also sparsifies the trajectory tree, reducing the computation time of many planner subroutines. 

\begin{figure}
\centering
\includegraphics[scale=0.45]{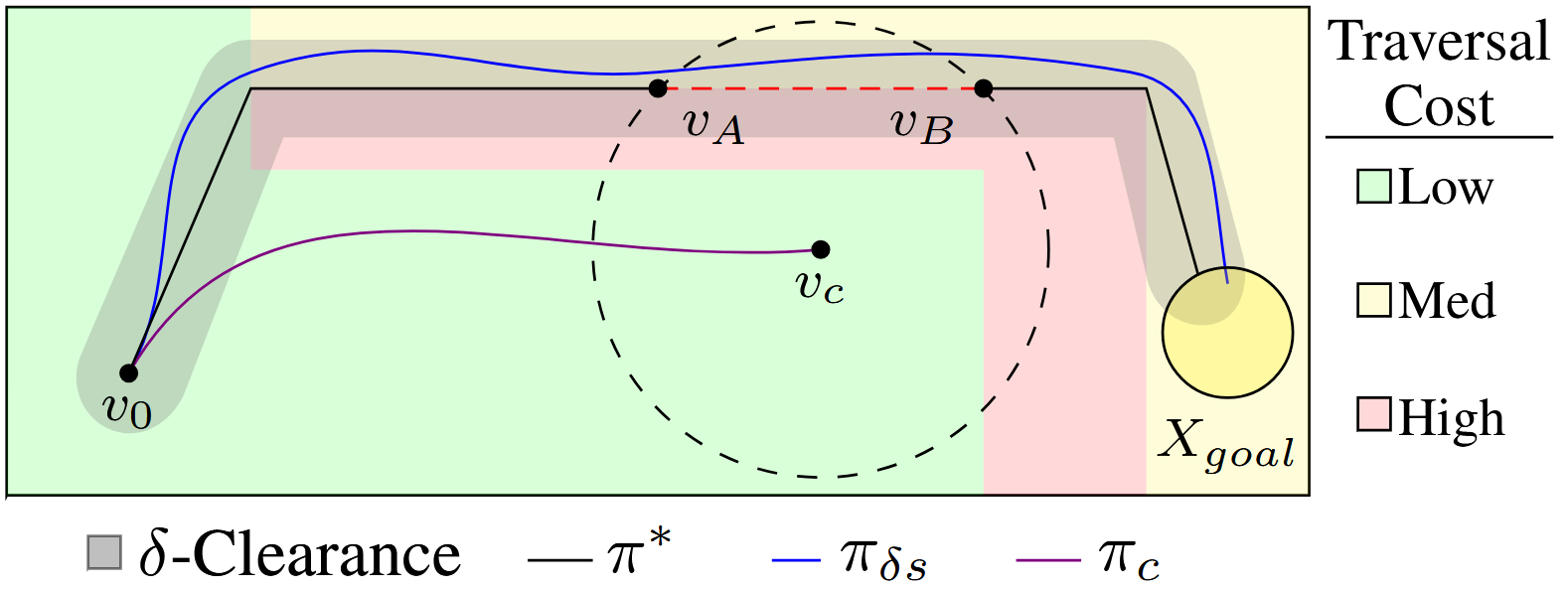}
\caption{While finding $\delta$-similar paths (blue) are traditionally a method of guaranteeing near-optimal algorithm properties, it is possible for certain non-$\delta$-similar paths (purple) to prevent the sampling of $\delta$-similar paths. Trajectories near $v_A \rightarrow v_B$ (dashed red) are invariably pruned due to the existence of $v_C$}
\label{fig:introfig}
\end{figure}

Many such sparse algorithms base proofs of near-optimality on the seminal work of the SST algorithm \cite{Littlefield2018EfficientRegions,Perrault2025Kino-PAX:Planner}.
In \cite{Li2016AsymptoticallyPlanning}, solution paths that are exist within a distance of at most $\delta$ from an optimal solution trajectory $\pi^*$ are shown to be near-optimal solution paths.
An example of such a trajectory is the blue trajectory in Figure \ref{fig:introfig}, $\pi_{\delta s}$.
The proof of the ANO property of SST implicitly assumes that $\delta$-similar trajectory segments will always be added to the trajectory tree once sampled.
This paper shows that this assumption is not true in general and that low-cost non-$\delta$-similar nodes can permanently dominate regions near the optimal solution path, such as the segment of $\pi^*$ in dotted red in Figure \ref{fig:introfig}.
This effect, referred to as the \textit{crowding out} of $\delta$-similar trajectories, prevents the retention of $\delta$-similar nodes.
The proof as originally structured therefore cannot conclude that $\delta$-similar nodes will be inductively added to the tree and the ANO property cannot be established without additional logic.

This work provides a new proof that shows that $\delta$-similar trajectories are not always guaranteed to be sampled, but that algorithmic near-optimality can still be achieved. 
By tracking the existence of near-optimal nodes directly rather than relying on the existence of $\delta$-similar nodes, the induction argument is repaired without requiring the stronger and generally unattainable claim of sampling $\delta$-similar solution trajectories.
As a result, it is shown that non-$\delta$-similar trajectories can be near-optimal.
A representative example is provided that illustrates the crowding-out effect, the guaranteed rejection of $\delta$-similar trajectory segments, and non-$\delta$-similar near-optimal solution trajectories.


\section{Problem Formulation}
This section introduces the motion planning problem and the main assumptions and definitions involved in proving near-optimality.

\subsection{System Assumptions} 
Consider kinodynamic systems described by 
\begin{equation}
\dot x(t) = f(x(t),u(t)), x(t) \in \mathbb{X}, u(t) \in \mathbb{U}
\label{eq:dynamics}
\end{equation}
where $\mathbb{X}$ is the state space with admissible space $\mathbb{X}_a \subseteq\mathbb{X}$ and $\mathbb{U}$ is the control space.
For the following analysis, the same set of system assumptions will be used as in \cite{Li2016AsymptoticallyPlanning} Section 3.
These assumptions include Chow's Condition \cite{Chow1940UberOrdnung}, a bounded second derivative, and Lipschitz continuity of the dynamics in both $x$ and $u$.
Additionally, the cost function used is assumed to be Lipschitz continuous, additive, monotonic, and non-degenerate.

\subsection{Definitions}

\begin{definition}[Trajectory]
A trajectory $\pi$ is a mapping $\pi(t):[0,t_\pi]\rightarrow\mathbb{X}_a$ generated by integrating Equation \ref{eq:dynamics} with a control function $u(t)$.
\end{definition}
Trajectory segments that are sampled and not pruned are added to a trajectory tree $G$. 
The endpoints of such trajectories are vertices on the tree where vertex $i$ is represented as $v_i$.
The location of $v_i$ in the state space is represented as $x_i$.

One common method of generating large, complex solution trajectories is to sample smaller, feasible trajectory segments using randomized control functions and concatenating them to form larger, piecewise trajectories.
Trajectory segments are sampled with a propagation time $t_{prop} \in (0,T_{prop})$. 
Selecting an appropriate value for $T_{prop}$ requires some domain specific knowledge. 
Excessively small values will generate trajectory segments that do not exit localized sparsity neighborhoods, while excessively large values produce highly collision-prone trajectory segments. 

\begin{definition}[Dynamic Clearance, Lemma 6 in \cite{Li2016AsymptoticallyPlanning}]
A path $\pi(t)$ has dynamic clearance $\delta$ if, for two points $x_i$ and $x_{i+1} \in \pi$,  $\forall x_i' \in \mathcal{B}_\delta(x_i)$ and $\forall x_{i+1}' \in \mathcal{B}_\delta(x_{i+1})$ there exists a trajectory $\pi'$ such that $\pi'(0) = x_{i}'$ and $\pi'(t_{\pi'}) = x_{i+1}'$.
\label{def:dynclr}
\end{definition}

$\mathcal{B}_\delta(x_i)$ is defined as the closed ball of radius $\delta$ centered on $x_i$. 
Dynamic clearance is an essential property of problems solvable by sampling-based planners. 
Around any path with positive dynamic clearance, there exists a set of feasible paths which have a non-zero probability of being sampled.
As in \cite{Li2016AsymptoticallyPlanning}, it is assumed that in all problems addressed in this work there exists an optimal solution trajectory with non-zero dynamic clearance.
Notably, it is not possible to create feasible paths to or from every point in a covering ball that includes an obstacle.
Therefore, dynamic clearance will always be less than or equal to obstacle clearance, which is defined as the minimum distance from any point on a trajectory to any point in collision with an obstacle.
The analysis in \cite{Li2016AsymptoticallyPlanning} uses the concept of $\delta$-robustness, which is the minimum of dynamic clearance and obstacle clearance for a path.
As obstacle clearance cannot be less than dynamic clearance, dynamic clearance and $\delta$-robustness can be conceptually used interchangeably. 

\begin{definition}[Near-Optimal Kinodynamic $\delta$-Robust Motion Planning]
Given an initial state $x_0$, a goal region $X_{goal}$, system constraints including dynamic constraints in Equation \ref{eq:dynamics} and the admissible space $\mathbb{X}_a$, and that an optimal solution path with dynamic clearance $\delta \in \mathbb{R}_{>0}$ and cost $c^*$ exists, find a solution trajectory $\pi$ such that $\pi(0) = x_0$ and $\pi(t_\pi) \in X_{goal}$, and $\texttt{cost}(\pi) \leq h(c^*,\delta)$.
\label{def:problem}
\end{definition}
The function $h(c^*,\delta)$ bounds the difference in cost between a trajectory found by a planning algorithm and the cost of the optimal solution trajectory. 
In contrast, many feasible planning algorithms have an unbounded cost difference \cite{LaValle2001RandomizedPlanning, Hsu2002RandomizedObstacles}.
The near-optimal $\delta$-robust motion planning problem assumes that there exists an optimal trajectory with positive dynamic clearance.

\begin{definition}[Asymptotic Near-Optimality]
An algorithm is asymptotically near-optimal if the cost of the best path found by the algorithm approaches a value bounded by a function $h(c^*,\delta)$ as the number of algorithm iterations approaches infinity.
\end{definition}

Near optimality for a motion planning algorithm is a guarantee of probabilistic performance as the algorithm is allowed to run. 
For algorithms solving the problem in Definition \ref{def:problem}, the probability that an algorithm will sample a solution trajectory with a sufficiently low cost is one as the number of iterations goes to infinity. 

\begin{definition}[Near-Optimal Node]
A node $x$ is near-optimal if there exists a near-optimal solution trajectory $\pi_{NO}$ such that $x \in \pi_{NO}$.
\end{definition}
Near-optimal nodes have the potential to be propagated into near-optimal solution trajectories.
The creation of such nodes is essential to the generation of near-optimal solution trajectories.

\begin{definition}[$\delta$-Similar Trajectories, Definition 3 in \cite{Li2016AsymptoticallyPlanning}]
Trajectories $\pi$ and $\pi'$ are $\delta$-similar if, for a continuous non-decreasing scaling function $\sigma : [0,t_\pi] \rightarrow [0,t_{\pi'}]$, $\pi'(\sigma(t)) \in \mathcal{B}_\delta(\pi(t))$.
\label{def:dstraj}
\end{definition}

For trajectories to be $\delta$-similar they must exist within a scaled time-parameterized state-space ball of each other for the entire duration of the trajectory.
In this work, trajectories will be referred to as $\delta$-similar if they are $\delta$-similar to $\pi^*$. 

In \cite{Li2016AsymptoticallyPlanning}, paths that are $\delta$-similar to an optimal solution trajectory $\pi^*$ are shown to be $\delta$-robustly near-optimal. 
A $\delta$-similar trajectory therefore is a solution to the near-optimal $\delta$-robust motion planning problem.
Finding $\delta$-similar trajectories, however, is sufficient but not necessary for finding a near-optimal trajectory.

\begin{definition} [$\delta$-Similar Node]
A node $v$ is a $\delta$-similar node if its source trajectory is $\delta$-similar to a subset of $\pi^*$.
\end{definition}

The source trajectory for $v$ is the trajectory originating at $x_0$ and terminating in $v$. 
If this trajectory is $\delta$-similar to a section of $\pi^*$, also from $x_0$ to some point near $v$, then $v$ can be considered a $\delta$-similar node.
Defining partial $\delta$-similarity for nodes and trajectories will be useful in examining individual algorithm iterations in the proof section later in this work. 

\begin{definition}[Covering Ball Sequence, Definition 14 in \cite{Li2016AsymptoticallyPlanning}]
For a trajectory $\pi(t)$ and a cost difference value $C_\Delta > 0$, the covering ball sequence $\mathbb{B}_\delta(\pi)$ is the set of $M+1$ hyper-balls $\{\mathcal{B}_\delta(x_0), \mathcal{B}_\delta(x_1),..., \mathcal{B}_\delta(x_M) \}$ where $x_i \in \pi$ and  $C(x_{i+1})-C(x_i) = C_\Delta$ for $i = 0,1,...,M-1$
\end{definition}
Covering balls exist in the state space and require a locally Euclidean state space so that distance can be formulated.
For a trajectory $\pi$ with dynamic clearance of $\delta$, any covering ball sequence $\mathbb{B}_\delta(\pi)$ will be collision free.

\subsection{Summary of the SST Algorithm}
\begin{algorithm}
\caption{\texttt{SST}}\label{alg:cap}
\begin{algorithmic}[1]
\State $G(\mathbb{V},\mathbb{E}) \gets \Call{Initialize\_Planner}{x_0}$
\For {$N\;iterations$}{}
\State $v_{selected} \gets$ \Call{BestFirstSelection}{}
\State $v_{new} \gets$ \Call{Dynamics\_Propagation}{$v_{selected}$}
\If{\Call{Collision\_Free}{$v_{selected}\rightarrow v_{new}$}}
\If{\Call{Is\_Node\_Locally\_Best}{$v_{new}$}}
\State Add $v_{new}$ to $G$
\State \Call{Prune\_Dominated\_Nodes}{$v_{new}, G$}
\EndIf
\EndIf
\EndFor
\State \Return $G$
\end{algorithmic}
\end{algorithm}

The SST algorithm is the canonical example of sparse sampling-based ANO algorithms for kinodynamic problems \cite{Li2016AsymptoticallyPlanning}. 
Such algorithms grow a sparse tree of feasible trajectories in the state space by an iterative three-step process: selecting a node, propagating it into more trajectories, and refining the tree.
The tree is initialized to the initial state and a witness, a data structure representing a state-space ball, is generated centered on it.
In SST, witnesses are generated with a fixed radius $\delta_s$, which is an algorithm hyperparameter set during initialization.
Algorithmically, each iteration consists of three subroutines: node selection, propagation, and graph revision. 
First, a node is selected from the existing trajectory tree using \texttt{BestFirstSelection}, which selects either the best cost node in the tree within a radius $\delta_{BN}$ from a randomly sampled state or the nearest node in the tree if the $\delta_{BN}$ ball contains no nodes.
Next \texttt{MonteCarloPropagation} applies a randomized control sample $u \in \mathbb{U}$ for a duration $t \in [0,T_{prop}]$ to forward integrate the system dynamics in Equation \ref{eq:dynamics} starting from the selected node.
The resulting state, $x_{new}$, and trajectory segment are then checked for collisions.

The final part of one iteration is the process of pruning the trajectory tree.
The $\texttt{LocallyBest}$ function calculates the nearest witness to $v_{new}$, measured as distance from $v_{new}$ to the center of the witness.
If $v_{new}$ lies outside all witnesses, a new witness is created centered on $x_{new}$ and $\texttt{LocallyBest}$ returns true.
Otherwise, $\texttt{LocallyBest}$ returns true if $v_{new}$ is of lower cost than the active node of the nearest witness and false if $v_{new}$ is of worse or equivalent cost. 
Therefore, the new node can either replace the existing representative or itself be pruned depending on relative cost.
Any branch of the tree that is entirely locally suboptimal is then pruned.
Only nodes which are the representatives of their respective witnesses (referred to as active nodes) can then be propagated in the next iteration.

\section{Achieving Asymptotic Near-Optimality}
Proving asymptotic near-optimality is accomplished via an induction proof. 
Let $\mathbb{B}_\delta(\pi^*)$ be the covering ball sequence constructed around the optimal trajectory $\pi^*$ which has dynamic clearance $\delta$. 
The induction proof will show that as the number of algorithm iterations approaches infinity, trajectory segments will almost surely be propagated from one covering ball into the next for each ball in the sequence.
While individual segments may not connect to each other or may be rejected entirely by the algorithm, the act of their sampling proves the existence of near-optimal nodes in the next ball in the sequence. 
By comparing such trajectory segments to possible $\delta$-similar trajectories, it can be proven that even non-$\delta$-similar trajectories can lead to near-optimal solutions. 

\subsection{Sampling Trajectories}
\begin{definition}[Crowding Out]
A witness $s^*$ is \textit{crowded out} if there exists a node $v_c\in s^*$ such that $\texttt{cost}(v_c) < \texttt{cost}(v_{\delta})$ for any node $v_\delta \in s^*$ that could be sampled via a $\delta$-similar trajectory. A covering ball $\mathcal{B}_\delta(x_i^*)$ is crowded out if every witness $s^* \cap \mathcal{B}_\delta(x_i^*) \neq \emptyset$ is crowded out.
\label{def:crowdout}
\end{definition}

When a witness is crowded out, the pruning function prioritizes keeping $v_c$ over other nodes that are similar in cost to $\pi^*$.
In fact, if elements of $\pi^*$ itself were to be sampled by the propagation function, they would also be immediately pruned.
The phenomenon of crowding out complicates the induction proof that is traditionally used to prove asymptotic near-optimality. 
In addition to witnesses, it will be shown later that entire covering balls can be crowded out.

\begin{theorem} [Sampling and Propagation]
If an active node exists in $\mathcal{B}_{\delta_{BN}}(x_i^*)$, the probability of sampling a trajectory segment that is $\delta$-similar to a subset of $\pi^*$ from $\mathcal{B}_\delta(x_i^*)$ to $\mathcal{B}_{\delta_{BN}}(x_{i+1}^*)$ is greater than zero. 
\label{thm:prop}
\end{theorem}

To sample a trajectory from $\mathcal{B}_\delta(x_i^*)$ to $\mathcal{B}_\delta(x_{i+1}^*)$, a node in $\mathcal{B}_\delta(x_i^*)$ must first be selected by $\texttt{BestFirstSelection}$.
If a node exists in $\mathcal{B}_{\delta_{BN}}(x_i^*)$ and the two algorithm hyperparameters $\delta_{BN}$ and $\delta_s$ are sized according to
\begin{equation}
\delta > \delta_{BN} + 2\delta_s,
\label{eq:ballsize}
\end{equation}
it is then shown in \cite{Li2016AsymptoticallyPlanning} that the probability of selecting an active node in $\mathcal{B}_\delta(x_i^*)$, $P_{select}^{(i)}$, is greater than zero. 
From Equation \ref{eq:ballsize}, note that $\mathcal{B}_{\delta_{BN}}(x_i^*) \subset\mathcal{B}_\delta(x_i^*)$.

Once an active node is selected, it is also shown in \cite{Li2016AsymptoticallyPlanning} that the probability of propagating it into $\mathcal{B}_{\delta_{BN}}(x_{i+1}^*)$, $P_{prop}^{(i)}$ is greater than zero.
Performing the selection-propagation process in a single iteration results in the generation of a trajectory segment from $\mathcal{B}_\delta(x_i^*)$ to $\mathcal{B}_\delta(x_{i+1}^*)$.
This event has probability of occurring of 
\begin{equation}
P_{select}^{(i)} \times P_{prop}^{(i)}>0
\label{eq:Pprop}
\end{equation}
and therefore will almost surely occur as the number of planner iterations approaches infinity.
\qed

Note that Theorem \ref{thm:prop} is concerned with the sampling and propagation of a $\delta$-similar trajectory segment, not its retention in the tree. 
Crowding out affects only the pruning step that follows propagation. 
Therefore, the probability of generating the segment in Equation \ref{eq:Pprop} is unaffected by crowding out and Theorem \ref{thm:prop} holds regardless of whether a covering ball is crowded out.

\begin{theorem} [Existence of Near-Optimal Nodes]
If a $\delta$-similar trajectory segment originating from a near-optimal node in $\mathcal{B}_\delta(x_i^*)$ and ending in $\mathcal{B}_\delta(x_{i+1}^*)$ is generated in iteration $j$, then at the end of iteration $j$, there will exist a near-optimal node in $G$ in $\mathcal{B}_\delta(x_{i+1}^*)$.
\label{thm:keep}
\end{theorem}

If a propagation has occurred terminating in node $v_{new} \in \mathcal{B}_\delta(x_{i+1}^*)$, four possible cases can occur as the result of the pruning process: 
\begin{enumerate}
\item $v_{new}$ lands outside all existing witnesses. A witness is created centered on $x_{new}$ and $v_{new}$ is made the representative of this witness.
\item $v_{new}$ lands in an existing witness and is of better cost than the representative node of that witness, replacing it as the active representative node.
\item $v_{new}$ lands in an existing witness and is of worse cost than the existing witness, a $\delta$-similar node. $v_{new}$ is pruned in favor of this preexisting node.
\item $v_{new}$ lands in an existing witness and is of worse cost than the existing witness, a \textit{non}-$\delta$-similar node, $v_c$. $v_{new}$ is pruned in favor of this preexisting node.
\end{enumerate}

In the first and second cases where $v_{new}$ is added to $G$, $v_{new}$ must be a near-optimal node as the addition of a $\delta$-similar trajectory segment incurs the addition of a bounded cost increment. 
Then it is clear that a near-optimal node now exists in the next covering ball for these two cases.
In the third case, this conclusion is also clearly true.

The fourth case, however, is more challenging to analyze. 
A $\delta$-similar node, $v_{new}$, is pruned from witness $s^*$, where $s^* \cap \mathcal{B}_\delta(x_{i+1}) \neq \emptyset$, in favor of a non-$\delta$-similar node, $v_c$.
This event can only occur if the $\texttt{cost}(v_{new}) > \texttt{cost}(v_c)$.
If $v_c$ has a cost that is lower than \textit{every} possible $\delta$-similar node in $s^*$, including elements of $\pi^*$ itself, then $v_c$ \textit{crowds out} $s^*$, as in Definition \ref{def:crowdout}.

If witnesses covering the ball $\mathcal{B}_\delta(x_{i+1}^*)$ are all crowded out, then it becomes impossible to add a $\delta$-similar node in $\mathcal{B}_\delta(x_{i+1}^*)$ to the trajectory tree. 
In such a case, it is impossible to guarantee the sampling of a $\delta$-similar solution trajectory through inductive reasoning.

However, as $v_{new}\in \mathcal{B}_\delta(x_{i+1}^*)$, there must exist a set of paths from $x_{new}$ to $X_{goal}$ that are $\delta$-similar to a subset of $\pi^*$.
Because of the dynamic clearance of $\pi^*$, there must exist a set of $\delta$-similar paths that pass through $x_c$. 
Note, that the trajectory ending in $v_c$ is not a $\delta$-similar trajectory.
If $v_c$ crowds out a witness such that a $\delta$-similar node is rejected in favor of $v_c$, then $v_c$ must have a cost less than that of $v_{new}$:

\begin{figure}
\centering
\includegraphics[scale=0.45]{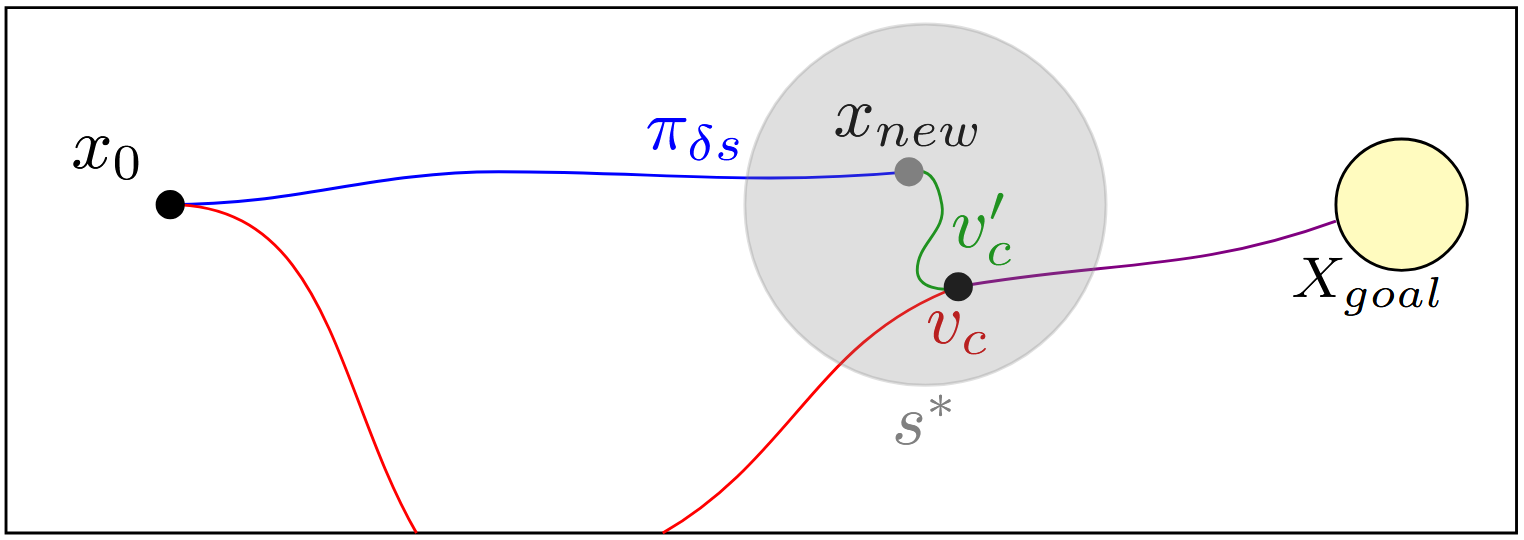}
\caption{A non-$\delta$-similar trajectory (red) crowds out a $\delta$-similar propagation (blue). Near-optimal solution trajectories are still able to be sampled through $v_c$, despite $v_c$ not being a $\delta$-similar node.}
\label{fig:ANO}
\end{figure}

\begin{equation}
\texttt{cost}(v_c) < \texttt{cost}(v_{new}).
\label{eq:ano1}
\end{equation}

\begin{figure*} [t]
\centering
\includegraphics[width=\linewidth]{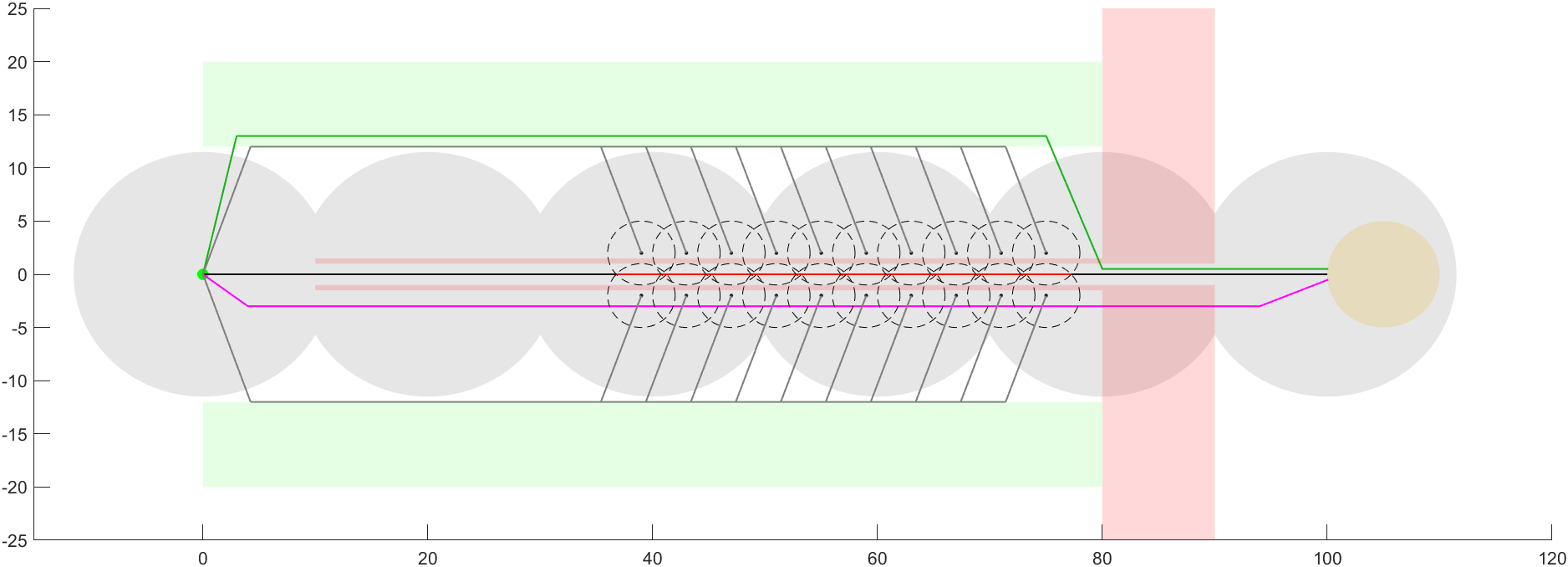}
\caption{
Environment consisting of regions of low traversal cost (green) and high traversal cost (red) with goal region (gold). The optimal solution path passes through the center of the environment. A set of valid covering balls is shown (shaded gray).}
\label{fig:env}
\end{figure*}

Also note that because both $x_{new} \in \mathcal{B}_\delta(x_{i+1})$ and $x_c \in \mathcal{B}_\delta(x_{i+1})$, a trajectory segment must exist that connects $x_{new}$ to $x_c$ (green in Figure \ref{fig:ANO}) that is $\delta$-similar to a zero-length subset of $\pi^*$ by Definition \ref{def:dynclr}.
This trajectory segment has positive cost.
Let the node that reaches state $x_c$ via a $\delta$-similar trajectory be $v_c'$.

The cost of $v_c'$ is given by
\begin{equation}
\texttt{cost}(v_c') = \texttt{cost}(v_{new}) + \texttt{cost}(v_{new}\rightarrow v_c').
\label{eq:ano2}
\end{equation}
Therefore, there must exist a $\delta$-similar trajectory from $v_0$ to $v_{new}$, to $v_c'$, to $X_{goal}$ (blue, green, purple in Figure \ref{fig:ANO}).
Let this path be $\pi_A$.
Then consider the cost of the solution trajectory $\pi_B = v_0 \rightarrow v_c \rightarrow X_{goal}$ (red and purple in Figure \ref{fig:ANO}).
For the purple trajectory connecting $v_c$ and $v_c'$, both located at state $x_c$, to $X_{goal}$, it must be possible to connect both nodes to the goal via the same trajectory, as both are located at state $x_c$.
Therefore, because cost is additive (Section 2.1), 
\begin{equation}
\texttt{cost}(v_c\rightarrow X_{goal}) = \texttt{cost}(v_c' \rightarrow X_{goal}).
\label{eq:ano3}
\end{equation}
Then, combining Equations \ref{eq:ano1}-\ref{eq:ano3}, the cost of the solution trajectory $\pi_B$ must be less than the cost of $\pi_A$:
\begin{equation}
\texttt{cost}(\pi_B) < \texttt{cost}(\pi_A).
\label{eq:ano4}
\end{equation}
As $\pi_A$ is a $\delta$-similar trajectory, its cost is shown to be of cost bounded by some function $h(c^*,\delta)$ in \cite{Li2016AsymptoticallyPlanning}, meaning that $\pi_A$ is a near-optimal trajectory: 
\begin{equation}
\texttt{cost}(\pi_A) \leq h(c^*,\delta).
\end{equation}
Because the cost of $\pi_B$ is shown to be less than the cost of $\pi_A$ in Equation \ref{eq:ano4}, it follows that
\begin{equation}
\texttt{cost}(\pi_B) < \texttt{cost}(\pi_A) \leq h(c^*,\delta).
\end{equation}
There then exists a node in $\mathcal{B}_\delta(x_{i+1}^*)$ with cost less than $h(c^*,\delta)$, meaning that a near-optimal node exists in $\mathcal{B}_\delta(x_{i+1}^*)$.

Therefore, in all four possible outcomes after the sampling of a $\delta$-similar trajectory segment from $\mathcal{B}_\delta(x_{i}^*)$ to $\mathcal{B}_\delta(x_{i+1}^*)$, there exists a near-optimal node in the next ball, proving Theorem \ref{thm:keep}.
\qed

The analysis of the fourth case after propagation and the effect of crowding out is the main contribution of this work.
Crowding out and the fourth case in general were not considered as a possible outcome in \cite{Li2016AsymptoticallyPlanning}, which implicitly assumes that $\delta$-similar trajectory segments are always added to the graph.
If the $\delta$-similar trajectory segments are always added, then $\delta$-similar solution trajectories are almost surely found by SST as only cases 1-3 apply.
Theorem \ref{thm:keep} allows an induction proof to be constructed that skips that intermediate claim and aims at asymptotic near-optimality directly.
This addition fills a notable gap in the foundational proof in \cite{Li2016AsymptoticallyPlanning}.

\subsection{Near-Optimality}
\begin{theorem}[Asymptotic Near-Optimality]
As the number of algorithm iterations approaches infinity, the SST algorithm almost surely samples a near-optimal trajectory.
\end{theorem}

The theorem is proved through induction.
For the inductive case, assume that there exists a near-optimal node in $\mathcal{B}_{\delta_{BN}}(x_i^*)$.
By Theorem \ref{thm:prop}, as the number of planner iterations goes to infinity, a $\delta$-similar trajectory segment from $\mathcal{B}_{\delta_{BN}}(x_i^*)$ to $\mathcal{B}_{\delta_{BN}}(x_{i+1}^*)$ is almost surely sampled. 
If one such trajectory is sampled at iteration $j$, at the end of iteration $j$ there is guaranteed to exist a near-optimal node in $\mathcal{B}_{\delta_{BN}}(x_{i+1}^*)$ by Theorem \ref{thm:keep}.
If this node is ever pruned at a later iteration, it must have been pruned in favor of a node with even lower cost, which is trivially also upper bounded by $h(c^*,\delta)$ and is therefore similarly near-optimal. 
Therefore, if the induction assumption is true for the $i$-th covering ball, it is almost surely true for the $(i+1)$-th covering ball as the number of iterations goes to infinity.
The root node $v_0$ exists at state $x_0$, which is equivalent to $x^*_0$, the center of $\mathcal{B}_{\delta_{BN}}(x_0^*)$.
With a cost of zero, $v_0$ is trivially near-optimal and satisfies the base case of the proof. 

It can therefore be concluded that, as the number of planner iterations approaches infinity, the algorithm will almost surely sample and keep a node within $\mathcal{B}_\delta(x_M^*)$ with a cost that is at least as good as a $\delta$-similar trajectory terminating in the same state. 
Such a node is a near-optimal node and is bounded in cost by the potential existence of a $\delta$-similar trajectory terminating in the same node.
The trajectory terminating in $\mathcal{B}_\delta(x_M^*)$ is therefore a near-optimal trajectory and is almost surely sampled by SST.
\qed

While $\delta$-similar trajectories may not be kept by the pruning function, the sampling of $\delta$-similar trajectory segments into the next covering ball in the sequence guarantees that that ball contains a near-optimal node.
In this way, the covering ball sequence is populated with near-optimal nodes, but not necessarily in order.
Logical progression through the entire sequence is necessary to confirm that populating nodes are near-optimal, but not necessarily to generate the nodes themselves.
Notably, the function $h(c^*,\delta)$ that bounds the cost of such non-$\delta$-similar near-optimal paths is the same function that bounds the cost of $\delta$-similar trajectories.

While this proof is specifically targeted at the SST algorithm, its logic applies to other similar sparse, forward-propagating, sampling-based algorithms.
The concept of pruning nodes based on localized regions of evaluation (in SST's case witnesses) is heavily used across this class of motion planning algorithm \cite{Littlefield2018EfficientRegions, Perrault2025Kino-PAX:Planner}.
These algorithms also share the three-part iteration structure that is integral to the induction proof. 

\section{Representative Example}
This section provides an example problem where the SST algorithm is not guaranteed to generate a $\delta$-similar solution trajectory due to crowding out, but non-$\delta$-similar near-optimal solution trajectories can still be found.

\subsection{Example Problem}

Consider the system of a kinematic point with unit speed with a controllable heading:
\begin{equation}
\label{eq:sys}
\mathbf{\dot{x}} =  
\begin{bmatrix} 
\cos(u) \\ 
\sin(u) \end{bmatrix}
\end{equation}
with control input $u \in \mathbb{U}$.
Let the state space be position in two-dimensions, $(x,y)$. 
Control inputs are assumed to take the form of piecewise constant control functions.

Let the cost of a trajectory be defined as the integral of the local traversability cost:
\begin{equation}
\texttt{cost}(\pi) = \int_0^{t_\pi} c_t(\pi(t)) dt
\end{equation}
where $c_t(x)$ is the cost per second incurred from traversing state $x$.

The environment in Figure \ref{fig:env} consists of an obstacle free space with several regions of differing traversability costs, the starting state $x_0$ (green point), and the goal region $X_{goal}$ (gold circle).
Green and red regions represent areas of low and high traversability cost and incur a cost of 1 and 300 per unit distance respectively.
All other regions incur 3 per unit distance.
Cost regions have continuous transitions with a finite Lipschitz constant determined by the maximum rate of change of $c_t(x)$ at the region boundaries. 
The system, therefore, can transit any region in the environment, but with varying costs depending on route. 

From examining the environment, the optimal solution path $\pi^*$ can be seen to be the straight line path between $x_0$ and $X_{goal}$.
The cost of $\pi^*$ is 300, as $\pi^*$ traverses a length of 100 entirely in unshaded regions where $c_t = 3$.
A valid covering ball sequence is also drawn in Figure \ref{fig:env} with a $\delta$ value of 11.5 m (gray circles).
This $\delta$ allows for $\delta_s$ and $\delta_{BN}$ to be set at 3 m and 5 m respectively, values taken from an example problem in \cite{Li2016AsymptoticallyPlanning}.
It is important to note that higher $\delta$ values for $\pi^*$ in this example are valid, which would allow a corresponding increase in $\delta_s$. 
However, arbitrary upper bounds can be placed on $\delta$ with the addition of obstacles near the start point or the goal region.
Such obstacles are not pictured in Figure \ref{fig:env}.
Let the maximum propagation time be $T_{prop} = 30$.
This allows propagations that can pass through any individual covering ball in the sequence, allowing for trajectories to reach the next covering ball from any point in the previous ball.

Several potentially sampled trajectories are shown (black lines) as well as several witnesses created at the end points of these trajectories (dashed circles).
As these trajectories pass through green regions before reaching their end points, the cost-to-arrive at the termination points through those trajectories is lower than the cost-to-arrive within the witnesses through $\pi*$. 
The red highlighted section of $\pi^*$ is therefore crowded out by the set of shown witnesses.
Samples both of $\pi^*$ and sampled trajectories near $\pi^*$ will be rejected by $\texttt{LocallyBest}$.

Any $\delta$-similar trajectory terminating in the fourth covering ball, $\mathcal{B}_\delta(x_4^*)$, in the sequence will be rejected, even ones outside of the narrow central corridor. 
The fourth covering ball can be reached via $\pi^*$ with a minimum cost of 145.5, the cumulative cost along $\pi^*$ to the edge of the ball.
However, the nodes that crowd out $\mathcal{B}_\delta(x_4^*)$ are reached with costs in the range of $[113,134]$.
Every such node is only reachable with such a low cost via trajectories that are outside of the covering ball sequence.
It can therefore be seen that it is not possible to generate a $\delta$-similar trajectory into $\mathcal{B}_\delta(x_4^*)$ that will not immediately be pruned in favor of one of the existing witnesses.

\subsection{Near-Optimal Trajectories}
In Figure \ref{fig:env}, the magenta trajectory stays entirely within the provided covering ball sequence and progresses consistently in the direction of $\pi*$. 
By Definition \ref{def:dstraj}, the magenta trajectory is a $\delta$-similar trajectory and is therefore near-optimal and must be cost-bounded by some $h(c^*,\delta)$.
Passing through the large high cost region before the goal incurs a large cost and the total cost of the magenta trajectory is approximately 3274, 1092\% worse than the optimal solution cost. 
This cost is dominated by the cost of traveling a length of 10 through the red region where $c_t = 300$, which incurs a cost of 3000 itself. 
Therefore, for this problem the value of $h(c^*,\delta)$ must be at least 3274 for a $c^*=300$ and $\delta = 11.5$.

To demonstrate that it is possible to sample a near-optimal, yet non-$\delta$-similar trajectory, see the green trajectory in Figure \ref{fig:env}.
As it benefits from traversing most of the environment in the green region and minimizing time in the red region, the cost of the green trajectory can be shown to be approximately 392, 30.7\% worse than the cost of the optimal trajectory.
As it is known that $h(c^*,\delta)$ is at least 3274, the green trajectory has cost less than $h(c^*,\delta)$ and therefore is near-optimal despite not being $\delta$-similar to $\pi^*$. 

\subsection{Propagating Over Covering Balls}
It is important to acknowledge that increasing the maximum propagation time allows for propagation through the crowded-out region, ignoring the crowding out effect. 
However, for a given $T_{prop}$, it is always possible to create an adversarial environment where propagation through the crowded-out region is impossible. 
The potential solution of increasing $T_{prop}$ is also highly impractical, as $T_{prop}$ would have to scale with the size of crowded-out regions, which are not initially known or even bounded in size.
Increasing $T_{prop}$ also reduces the probability that any given trajectory segment is collision-free, reducing performance. 
Carried far enough, as $T_{prop}$ approaches levels greater than or equal to $t_{\pi^*}$, the algorithm reduces to attempting to solve the entire planning problem with a single propagation.
Such behavior is exactly what sampling-based iterative motion planning is meant to avoid. 

\section{Conclusions}
Many forward-propagating sparse SBMP algorithms prove asymptotic near-optimality through the claim that a solution that is $\delta$-similar to the optimal solution path will almost surely be found given sufficient iterations. 
However, in some situations, $\delta$-similar trajectories are guaranteed to be pruned from the tree. 
This work provides a refined proof of asymptotic near-optimality without claiming guarantees of sampling $\delta$-similar paths.
The new proof applies generally to similar sparse algorithms that utilize the claim of almost-surely sampling $\delta$-similar trajectories.
An example planning problem is provided where $\delta$-similar trajectories are guaranteed to be rejected from certain regions near the optimal path.
In such an environment, it is possible to sample near-optimal paths, despite the guaranteed pruning of $\delta$-similar paths. 

\bibliographystyle{IEEEtran}
\bibliography{references}

\end{document}